# TRANSFORMER FOR IMAGE QUALITY ASSESSMENT

*Junyong You[1], Jari Korhonen[2]*

1. NORCE Norwegian Research Centre, Bergen, Norway (junyong.you@ieee.org);
2. Shenzhen University, Shenzhen, China (jari.t.korhonen@ieee.org)

**ABSTRACT**

Transformer has become the new standard method in natural language processing (NLP), and it also attracts research interests in computer vision area. In this paper we investigate the application of Transformer in Image Quality (TRIQ) assessment. Following the original Transformer encoder employed in Vision Transformer (ViT), we propose an architecture of using a shallow Transformer encoder on the top of a feature map extracted by convolution neural networks (CNN). Adaptive positional embedding is employed in the Transformer encoder to handle images with arbitrary resolutions. Different settings of Transformer architectures have been investigated on publicly available image quality databases. We have found that the proposed TRIQ architecture achieves outstanding performance. The implementation of TRIQ is published on Github (https://github.com/junyongyou/triq).

## 1. INTRODUCTION

Transformer [1] was first proposed to replace recurrent neural networks (RNN), e.g., long short-term memory (LSTM), in an encoder-decoder architecture for machine translation tasks. It is purely based on attention mechanisms, including self-attention and encoder-decoder attention, which can overcome intrinsic shortages of RNN. Lately, Transformer has been adopted in computer vision (CV) tasks, e.g., object detection (DETR) [2] and image recognition (ViT) [3].

Following the original idea in Transformer, DETR [2] employs the encoder-decoder architecture to predict either a detection (object class and bounding box) or a background class. Relevant image features are first learned from 2D convolution neural networks (CNN), and then fed into the encoder together with positional embedding. The decoder takes a small fixed number of learned positional embedding and attends to the encoder output for object detection. In ViT for image recognition [3], only the Transformer encoder is employed. An input image is first divided into small patches (e.g., 16×16 pixels), which are then flatten and projected onto patch embedding by a linear projection layer. The linear projection works similarly to the word embedding in the original Transformer and can also match the image patch dimension to the model dimension of Transformer encoder. Subsequently, an extra embedding, similar to the [class] token in BERT [4], is added in the beginning of the projected embeddings. This extra embedding is expected to represent the aggregated information for image representation on the whole set of image patches without being biased to any particular patches. Learnable positional embedding is added to the embeddings to retain positional information. The encoder architecture is the same as the original Transformer. This is also a common approach of applying Transformer in CV tasks, allowing the latest developments of Transformer to be directly applied. Consequently, a multi-layer perceptron (MLP) head is added on the top of the Transformer encoder for image recognition. A hybrid approach to apply Transformer encoder on CNN features has also been suggested in the ViT paper [3].

Image quality assessment (IQA) can be considered essentially as a recognition task, i.e., recognizing the quality level of an image. Therefore, we attempt to investigate how to apply Transformer in the IQA task. Existing deep learning driven IQA models are mainly based on CNN architectures. A typical example is to use CNN as a feature extractor and MLP on the top to predict image quality. Due to hardware restrictions, earlier works applied CNNs on small image patches, e.g., AlexNet in [5] and VGG in [6], and then combined the quality predictions from the patches into a single quality indicator. Later, larger image patches (e.g., 224×224 pixels) have been used [7]. However, patch-based models often assume that image patches share the same quality level as their original full image, when training the models. Such assumption might not be reliable, as image quality can be spatially varying or viewers might not pay similar attention to different image regions. Thus, an end-to-end approach predicting quality from the whole image is more desired in IQA.

Considering that image quality can be affected by spatial saliency distribution, Yang *et al.* [8] proposed an end-to-end multi-task network (SGDNet) to predict saliency map and image quality jointly. Image saliency map can also be used as input to derive more perceptually consistent features for quality prediction. Inspired by the mechanism of hierarchical representations in visual system [9] and object detection tasks [10], Wu *et al.* [11] proposed a cascaded architecture (CaHDC) to derive features at different scales, and then combined them into quality prediction by a side pooling net. In [12], Hosu *et al.* first built a large scale image quality database (KonIQ-10k). A simple yet effective architecture has then been proposed consisting of a base CNN network (Inception-Resnet V2) followed by MLP.

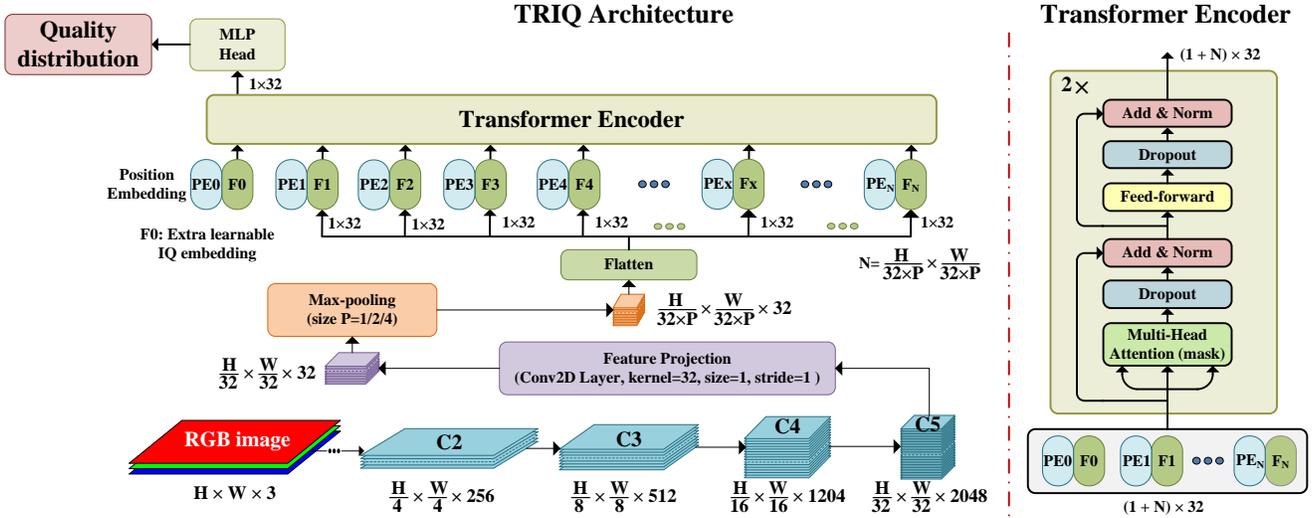

Fig.1. Architecture of TRIQ and Transformer encoder (the numbers indicate actual output shape of layers/blocks in our experiments).

IQA has a particular characteristic that distinguishes it from other CV tasks (e.g., object detection or image recognition): image quality can be significantly influenced by resolution. In other words, image quality can be potentially affected by resizing, e.g., significantly down-sampling images might degrade their perceived quality. Even though image resizing is widely employed in deep learning models for other CV tasks, it should be avoided in IQA models. In principle, Transformer can accept inputs with varying lengths (e.g., sentences with varying numbers of words). Thus, it is also possible to adapt the Transformer encoder to images with different resolutions to build a generic IQA model.

The remainder of the paper is organized as follows. Section 2 presents the Transformer encoder for Image Quality (TRIQ) assessment. Detailed experiments are discussed in Section 3, and Section 4 draws the concluding remarks.

## 2. TRIQ: TRANSFORMER FOR IMAGE QUALITY ASSESSMENT

ViT [3] contains two approaches: image patching-based and hybrid. The former is to divide an image into patches with fixed size (e.g., 32×32 pixels), and the latter uses a feature map produced by other CNN base networks (e.g., ResNet).

The flattened image patches or feature map will then be fed into the Transformer encoder. In order to better unify the patching-based and hybrid approaches, a 2D convolutional layer (Conv2D-projection) is employed for feature projection. The number of kernels in the projection layer is set to the model dimension (*D*) of the Transformer encoder, as Transformer uses constant latent vector size *D*. For the image patching approach, the kernel size and stride of the Conv2D-projection layer are both set to the same as the patch size. For example, if a patch size 32×32 pixels is used, the projected features have a shape of [H/32, W/32, *D*], where *H*, *W* denote the image height and width. On the other hand, if the hybrid approach is used, the activation map of the final block of the CNN is employed as the feature map. For example, if ResNet50 is used as the base network, it produces the feature map with a shape of [H/32, W/32, 2048], where 2048 is the channel dimension. Subsequently, the kernel size and stride in the Conv2D-projection layer are set to 1. Consequently, the projected features also have a shape [H/32, W/32, *D*].

The proposed TRIQ architecture follows the hybrid approach, as illustrated in Fig. 1. ResNet50 was chosen as the base network. It should be noted that other CNN architectures can also be used.

ViT for image recognition can take constant input size with the help of image resizing. However, as explained earlier, image resizing should be avoided in IQA. Therefore, TRIQ needs to handle images with different resolutions. This mainly concerns the positional embedding representing positional information of image patches or CNN features for image quality perception. To solve this issue, we define the positional embedding with sufficient length to cover the maximal image resolution in our datasets, which is then truncated for individual smaller images.

Furthermore, when an input image has large resolution, the number of image patches or CNN features (e.g., H/32 × W/32) can also be large. This causes two potential issues. The first is that a lot of memory is required to run the Transformer encoder on an image with a very high resolution. Second, a large number of patches/ features will also cause difficulty to the Transformer encoder to capture long-term dependence across the patches/ features. Therefore, max-pooling is performed, depending on the input image resolution. For TRIQ with hybrid Transformer, max-pooling is applied to the CNN features before feeding them into the Conv2D-projection layer. Whereas, for TRIQ with patching approach, image patches are first fed into the Conv2D-projection layer, and then max-pooling is performed. The pooling size is adaptively determined based on the resolution of the input image. Thus, the projected features from the Conv2D-projection layer have a shape [H/(32×P), W/(32×P), D], where P is the max-pooling size.

Subsequently, the two spatial dimensions of the pooled features are first flattened to shape [N=(H×W)/(32×32×P×P), D], and then the learnable positional embeddings (PE) are added. As explained in ViT and BERT [4], an extra embedding ($F_0$) is appended in front of the projected features (F), which is also added by a positional embedding ($PE_0$). Eq. (1) roughly explains the projection, max-pooling and positional embedding, where FM denotes the feature map.

$$\begin{cases} F_j = Max[Conv2D\_proj(FM)], & j = 1, \cdots N \\ Z_0 = [F_0 + PE_0; \cdots F_N + PE_N], & PE_j \in R^{(1+N) \times D} \end{cases} \quad (1)$$

The Transformer encoder, specified essentially by four key hyper-parameters: L (number of layers), D (model dimension), H (number of heads), and $d_{ff}$ (dimension of the feed-forward network), consists of an alternative number of encoder layers. As shown in Fig. 1, each encoder layer contains two sublayers: the first is for multi-head attention (MHA), and the second contains position-wise feed-forward (FF) layer. Layer normalization (LN) is performed on the residual connection in the two sublayers. The Transformer encoder, as briefly explained in Eq. (2), produces an output of shape (1+N)×D.

$$\begin{cases} Z'_l = LN(MHA(Z_{l-1}) + Z_{l-1}) \\ Z_l = LN(FF(Z'_l) + Z'_l) \end{cases} \quad l = 1, \cdots L \quad (2)$$

Finally, the first vector of output ($Z_L[0]$), that is supposed to contain aggregated information from the Transformer encoder for image quality perception, is fed into an MLP head. The MLP head consisting of two fully connected (FC) layers and a dropout layer in between predicts the perceived image quality. The first FC layer uses $d_{ff}$ filters with GELU activation, as suggested in BERT [4] and ViT [3]. Earlier studies have demonstrated that predicting quality distribution over grade to vote (e.g., 1=bad, 2=poor, 3=fair, 4= good, 5=excellent) often provide more robust results than single MOS values [13]. Thus, we use five filters in the last FC layer with Softmax activation to predict the quality distribution, i.e., normalized probabilities over the five quality grades. Consequently, cross entropy is chosen as the loss function to measure the distance between the predicted image quality distribution and the ground-truth distribution. Assuming p(x) denotes the normalized probability predicted by TRIQ, a single MOS value can be easily derived as defined in Eq. (3).

$$MOS = \sum_{x \in \{1,2,3,4,5\}} x \cdot p(x) \quad (3)$$

## 3. EXPERIMENTS

### 3.1. Experiment settings

Two large-scale image quality databases have been employed in the experiments, namely KonIQ-10k [12] and LIVE-wild databases [14]. Both the MOS values and quality score distribution in the KonIQ-10k database were published by the authors. The authors of LIVE-wild database released standard deviations of quality scores, and the score distribution can be derived by assuming the scores follow a truncated Gaussian distribution. The two databases share many similarities, e.g., same methodology of crowdsourcing, and natural image content with non-specific distortions. However, it is inappropriate to combine IQA databases from different experiments directly. The main concern is calibration of rating scales, e.g., identical image content might be assigned with dissimilar quality levels in different experiments. We have conducted a small-scale experiment using five participants to compare 20 image pairs to align the KonIQ-10k and LIVE-wild databases. An image pair consisting of an image from KonIQ-10k and another from LIVE-wild with the same rating scales and the participants were asked to compare them and to judge if they represented a similar quality level.

The calibration experiment suggested that almost all the randomly selected 20 image pairs share similar quality levels. Thus, the two databases were combined in this work. Subsequently, we split the two databases into training and testing sets randomly according to both MOS values and image complexity, measured by spatial information (SI) as defined in the ITU Recommendation [15]. All images were first roughly classified into two complexity categories: high and low SI, and then the images in each category are further divided into five quality categories based on their MOS values. As the number of images in the KonIQ-10k database is much higher LIVE-wild, we randomly chose 85% of the images in KonIQ-10k and 50% of the images in LIVE-wild from each quality category in each complexity category as training images, and the rest of the images as testing images. In addition, the authors of KonIQ-10k database also used half-sized images in the database with the quality scores voted on the original full resolution images to develop the Koncept512 model [12]. We followed this approach to include the half-sized images, named as KonIQ-half-sized database, in our experiment. Consequently, a combined database was generated consisting of KonIQ-10k, KonIQ-half-sized, and LIVE-wild databases, and the contained training and testing sets were inherited from the original splits of the individual databases.

Three state-of-the-art IQA models, namely SGDNet [8], CaHDC [11] and Koncept512 [12], were included in our experiments as a comparison point. These models can be adjusted to accept images with varying resolutions by not specifying input size and using global pooling before the prediction layers. The models were retrained on our training and testing sets using the original implementations and training strategies published by the authors, respectively. It should be noted that SGDNet requires a saliency map to predict image quality, which is unavailable in the LIVE-wild database. Thus, SGDNet was not included in the experiments involving the LIVE-wild database.

In TRIQ, the output from the last residual block is chosen as the feature map. Subsequently, a shallow architecture for the Transformer encoder [$L$=2, $D$=32, $H$=8, $d_{ff}$=64] has been employed, which obtained the highest prediction accuracy in our experiments. Other settings for the Transformer encoder have also been tested, while we found that they both produce worse results than the above one.

The official implementations of ViT contain a patching based approach using two patch sizes and a hybrid approach based on Resnet50 V2 as base network. Model variants based on configurations for BERT have been proposed. Accordingly, the pretrained weights on large-scale databases have also been released [16]. In our experiments, we have based on two base ViT models (ViT-B_16 and R50+ViT-B_16 [16]) to predict image quality. The published weights pretrained on ImageNet21k have been employed for applying transfer learning for IQA in our experiments. The two models are named as ViT-IQA and R50+ViT-IQA, respectively.

Subsequently, TRIQ, ViT-IQA and R50+ViT-IQA were trained on the training set using Adam as optimizer. A learning rate scheduler with linear warm-up and cosine decay was applied. A base learning rate 5e-5 was used for pretraining and the best model weights were stored when they produced the highest Pearson correlation (PLCC) between the predicted MOS values and ground-truth on the testing sets. Subsequently, model finetuning was performed with a smaller learning rate 1e-6, based on the best weights produced in the pretraining phase.

### 3.2. Evaluation results and discussions

Three criteria, namely PLCC, Spearman rank-order correlation (SROCC), and root mean squared error (RMSE) between MOS predictions and ground-truth values, have been employed to evaluate individual IQA models in different scenarios. All the studied models were trained on the combined database and separate databases, respectively.

Table I. Evaluation results on the Combined testing set

| Models | PLCC↑ | SROCC↑ | RMSE↓ |
|---|---|---|---|
| CaHDC | 0.625 | 0.628 | 0.473 |
| Koncept512 | 0.719 | 0.755 | 0.601 |
| ViT-IQA | 0.775 | 0.764 | 0.388 |
| R50+ViT-IQA | 0.858 | 0.835 | 0.304 |
| TRIQ | *0.884* | *0.868* | *0.280* |

Table II. Evaluation results on the entire SPAQ database

| Models | PLCC↑ | SROCC↑ | RMSE↓ |
|---|---|---|---|
| CaHDC | 0.647 | 0.656 | 0.720 |
| Koncept512 | 0.817 | 0.825 | 0.607 |
| ViT-IQA | 0.754 | 0.762 | 0.585 |
| R50+ViT-IQA | 0.845 | 0.845 | 0.534 |
| TRIQ | *0.848* | *0.857* | *0.480* |

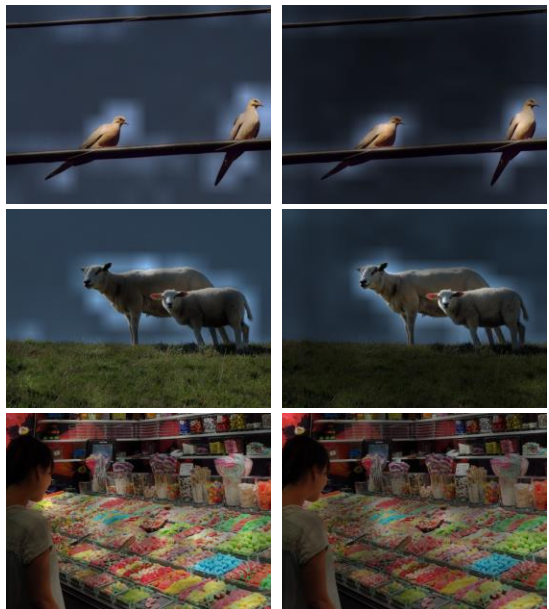

Fig. 2. Distribution of average attention weights over multi-heads (left: TRIQ model; right: ViT model for image recognition)

In the first experiment, the models were trained on the training set of the combined database, and then evaluated on the testing set. Table I reports the evaluation results, which demonstrate that TRIQ outperforms significantly other IQA models. Even though patching-based ViT has achieved promising performance in image recognition with the help of pretrain on large-scale databases, it does not beat TRIQ in IQA. We suspect the main reason is that relatively small-scale databases are used for IQA in this work, and they cannot take the full advantage of large-scale pretraining. The hybrid approach of ViT, including the proposed TRIQ, has shown outstanding performance in IQA. We assume this is because the inductive capability of CNN architectures can more appropriately capture image quality features than patches purely. As the employed datasets are all in small scale, a shallow architecture of Transformer encoder used in TRIQ, rather than a deeper one in R50+ViT-IQA, is sufficient for capturing the characteristics of CNN-derived features for image quality perception.

Table III. IQA models trained on **Combined** training set
and evaluation on the separate testing sets of KonIQ-10k & KonIQ-half-sized and LIVE-wild databases

| Models | Testing set of KonIQ-10k | | | Testing set of KonIQ-half-sized | | | Testing set of LIVE-wild database | | |
|---|---|---|---|---|---|---|---|---|---|
| | PLCC↑ | SROCC↑ | RMSE↓ | PLCC↑ | SROCC↑ | RMSE↓ | PLCC↑ | SROCC↑ | RMSE↓ |
| CaHDC | 0.702 | 0.721 | 0.442 | 0.731 | 0.701 | 0.386 | 0.551 | 0.542 | 0.693 |
| Koncept512 | 0.789 | 0.828 | 0.570 | 0.749 | 0.788 | 0.637 | 0.693 | 0.704 | 0.592 |
| TRIQ | ***0.923*** | ***0.909*** | ***0.213*** | ***0.895*** | ***0.868*** | ***0.249*** | ***0.826*** | ***0.812*** | ***0.449*** |

Table IV. IQA models trained on **KonIQ-10k** training set
and evaluation on the testing sets of KonIQ-10k & KonIQ-half-sized and entire LIVE-wild database

| Models | Testing set of KonIQ-10k | | | Testing set of KonIQ-half-sized | | | Entire LIVE-wild database | | |
|---|---|---|---|---|---|---|---|---|---|
| | PLCC↑ | SROCC↑ | RMSE↓ | PLCC↑ | SROCC↑ | RMSE↓ | PLCC↑ | SROCC↑ | RMSE↓ |
| SGDNet | 0.840 | 0.819 | 0.300 | 0.720 | 0.664 | 0.396 | — | — | — |
| CaHDC | 0.750 | 0.729 | 0.368 | 0.611 | 0.544 | 0.502 | 0.414 | 0.384 | 0.746 |
| Koncept512 | 0.854 | 0.854 | 0.293 | 0.745 | 0.722 | 0.374 | 0.738 | 0.716 | 0.591 |
| TRIQ | ***0.925*** | ***0.907*** | ***0.212*** | ***0.837*** | ***0.818*** | ***0.349*** | ***0.805*** | ***0.781*** | ***0.503*** |

Table V. IQA models trained on **KonIQ-half-sized** training set
and evaluation on the testing sets of KonIQ-10k & KonIQ-half-sized and entire LIVE-wild database

| Models | Testing set of KonIQ-10k | | | Testing set of KonIQ-half-sized | | | Entire LIVE-wild database | | |
|---|---|---|---|---|---|---|---|---|---|
| | PLCC↑ | SROCC↑ | RMSE↓ | PLCC↑ | SROCC↑ | RMSE↓ | PLCC↑ | SROCC↑ | RMSE↓ |
| SGDNet | 0.772 | 0.777 | 0.431 | 0.823 | 0.798 | 0.320 | — | — | — |
| CaHDC | 0.627 | 0.658 | 0.458 | 0.763 | 0.723 | 0.364 | 0.578 | 0.575 | 0.686 |
| Koncept512 | 0.869 | 0.876 | 0.575 | 0.923 | 0.901 | 0.215 | ***0.819*** | ***0.806*** | 0.681 |
| TRIQ | ***0.893*** | ***0.882*** | ***0.327*** | ***0.925*** | ***0.905*** | ***0.211*** | 0.800 | 0.779 | ***0.493*** |

Table VI. IQA models trained on **KonIQ-10k** & **KonIQ-half-sized** training sets
and evaluation on the testing sets of KonIQ-10k & KonIQ-half-sized and entire LIVE-wild database

| Models | Combined Testing set of KonIQ-10k & KonIQ-half-sized | | | Testing set of KonIQ-10k | | | Testing set of KonIQ-half-sized | | | Entire LIVE-wild database | | |
|---|---|---|---|---|---|---|---|---|---|---|---|---|
| | PLCC | SROCC | RMSE | PLCC | SROCC | RMSE | PLCC | SROCC | RMSE | PLCC | SROCC | RMSE |
| SGDNet | 0.813 | 0.782 | 0.322 | 0.860 | 0.843 | 0.284 | 0.766 | 0.721 | 0.355 | — | — | — |
| CaHDC | 0.683 | 0.669 | 0.411 | 0.693 | 0.723 | 0.436 | 0.750 | 0.718 | 0.385 | 0.566 | 0.575 | 0.687 |
| Koncept512 | 0.807 | 0.817 | 0.342 | 0.871 | 0.870 | 0.281 | 0.785 | 0.809 | 0.285 | 0.708 | 0.722 | 0.613 |
| TRIQ | ***0.900*** | ***0.880*** | ***0.244*** | ***0.919*** | ***0.906*** | ***0.219*** | ***0.889*** | ***0.861*** | ***0.266*** | ***0.845*** | ***0.784*** | ***0.500*** |

Table II reports the results of the models trained on the combined training set and then evaluated on the SPAQ database [17]. Interested readers can refer to [17] for the details about the SPAQ database. We tested the IQA models on the entire SPAQ database. TRIQ still shows promising performance across databases.

In addition, the trained models were also tested on individual testing sets from respective databases, and Table III gives the evaluation results. This demonstrates promising performance of TRIQ as a generic IQA model for diverse image contents and resolutions, compared with other state-of-the-art models.

In order to evaluate the model performance across resolutions and databases, we have also trained the models on KonIQ-10k, KonIQ-half-sized databases, and their combination, respectively. The models were evaluated on both the testing sets of KonIQ-10k and KonIQ-half-sized databases and the entire LIVE-wild database. Tables IV, V and VI report the respective evaluation results. ViT-IQA and R50+ViT-IQA still perform worse than TRIQ in these experiments. According to the results, cross-resolution and cross-database clearly have a negative impact on the performance of IQA models, even though TRIQ shows promising potential also in this scenario. This issue will be investigated to develop a more generic IQA model in the future work.

Furthermore, as the self-attention mechanism is employed in the Transformer encoder, it is interesting to investigate how attention works in IQA. Thus, the attention weights learned during the training process are visualized by averaging the attention weights over different heads, and then applying the normalized weights as a mask on the input image. As comparison, the pretrained ViT-B16 weights [16] for image recognition were also used to present the attention distribution in recognition task. Fig. 2 shows a few randomly chosen images with the masked attention distribution. Compared with image recognition where attention is mainly allocated to semantic objects, the attention distribution is spreading more widely in IQA, rather than focusing on regions of interest. We assume this is because the voters (subjective participants or objective quality models) often need to concentrate on the whole image area to collect more information when assessing image quality.

## 4. CONCLUSIONS

A promising IQA model TRIQ has been proposed in this work. TRIQ takes the advantage of inductive capability of CNN architecture for quality feature derivation and Transformer encoder for aggregated representation of attention mechanism. TRIQ has demonstrated outstanding performance on two publicly available large-scale image quality databases, compared with other deep learning driven IQA models. TRIQ shows another promising example to exploit Transformer in computer vision tasks. Future work will focus on developing more generic IQA models that can adapt to diverse image contents, resolutions and distortion types.